# An automatic bad band preremoval algorithm for hyperspectral imagery

Luyan Ji, Xiurui Geng, Yongchao Zhao, Fuxiang Wang

**Abstract**

For most hyperspectral remote sensing applications, removing bad bands, such as water absorption bands, is a required preprocessing step. Currently, the commonly applied method is by visual inspection, which is very time-consuming and it is easy to overlook some noisy bands. In this study, we find an inherent connection between target detection algorithms and the corrupted band removal. As an example, for the matched filter (MF), which is the most widely used target detection method for hyperspectral data, we present an automatic MF-based algorithm for bad band identification. The MF detector is a filter vector, and the resulting filter output is the sum of all bands weighted by the MF coefficients. Therefore, we can identify bad bands only by using the MF filter vector itself, the absolute value of whose entry accounts for the importance of each band for the target detection. For a specific target of interest, the bands with small MF weights correspond to the noisy or bad ones. Based on this fact, we develop an automatic bad band preremoval algorithm by utilizing the average absolute value of MF weights for multiple targets within a scene. Experiments with three well known hyperspectral datasets show that our method can always identify the water absorption and other low signal-to-noise (SNR) bands that are usually chosen as bad bands manually.

Keywords: hyperspectral; matched filter; bad band preremoval

**1, Introduction**

Hyperspectral remote sensing data has been widely used for land cover and land use mapping (Martin et al. 1998; Hestir et al. 2008; Roger N. Clark et al. 1990), target detection (Manolakis and Shaw 2002; Stein et al. 2002; Manolakis, Siracusa, and Shaw 2001), vegetation biomass and crop yield prediction (Tian et al. 2011; Huang et al. 2007; Cho et al. 2007), etc. In all these fields, one common preprocessing step is to remove the bad or noisy bands, such as atmospheric water absorption bands, low signal-to-noise (SNR) bands, and those with error patches (Mianji and Ye 2011; Camps-Valls et al. 2014). The use of these low SNR bands will lead to a lower accuracy of a classifier, detector or predictor. Even in the field of band selection, which is usually regarded as a preprocessing procedure, the preremoval of bad bands is also required (Qian and He 2008; Kang et al. 2014; Sun, Geng, and Ji 2014).

Currently, the most commonly used way to remove these bad bands is by visual interpretation (Qian and He 2008). Noisy bands within the atmospheric water absorption ranges are relatively easy to detect based on the wavelength information. However bands not in this wavelength range are hard to find unless we traverse all bands. Since the number of hyperspectral images is usually very large, the traversal process is very time-consuming. Another possible band preremoval method is to use the band selection technique, which aims to find the most distinctive and informative bands. However, Sun's research shows that some band selection methods may choose water absorption or other low SNR bands as the final representative ones (Kang, Xiurui, and Luyan 2015). Therefore, many band selection methods also require the removal of those bad bands in advance (Qian and He 2008).

In this study, we try to bridge a link between the bad band removal technique and the target detection method, which seem to be completely unrelated. In the field of target detection, the matched filter (MF) is the most commonly used technique for hyperspectral imagery (Manolakis et al. 2009, 2009; DiPietro et al. 2010; Manolakis et al. 2001; Funk et al. 2001; Minet et al. 2011). The MF output is a grey image with high values in target regions and low values in background regions. It can be considered as the sum of all bands weighted by $L$ MF detector weights ($L$ is the total number of bands).

Therefore, the weight coefficients actually indicate the significance of each band in the role of distinguishing the target from the background. Motivated by this fact, we develop an automatic bad band preremoval method in this paper.

**2. Method**

**2.1 Background**

MF has been widely used in communications, signal processing and pattern recognition applications. It is usually derived by maximizing the cost function, which measures the distance between the means of two normal distributions in units of the common variance (Manolakis and Shaw 2002). Assume that we are given a finite set of observations $\mathbf{R} = [\mathbf{r}_1, \mathbf{r}_2, \cdots, \mathbf{r}_N]$, where $\mathbf{r}_i = (r_{i1}, r_{i2}, \cdots, r_{iL})^T$ for $1 \leq i \leq N$ is a sample pixel vector, $N$ is the total number of pixels, and $L$ is the number of bands. Suppose that the desired signature $\mathbf{d}$ is also known. The expression of an MF detector can be written as (Manolakis and Shaw 2002)

$$\mathbf{w}_{MF} = \kappa \mathbf{K}^{-1}(\mathbf{d} - \mathbf{m}), \tag{1}$$

where $\mathbf{w}_{MF} = [w_1, w_2, \cdots, w_L]$ is an $L$-dimensional vector, $\mathbf{m} = (1/N)\sum_{i=1}^{N} \mathbf{r}_i$ is the mean vector, $\mathbf{K} = (1/N)\left[\sum_{i=1}^{N}(\mathbf{r}_i - \mathbf{m})(\mathbf{r}_i - \mathbf{m})^T\right]$ is the covariance matrix, and $\kappa$ is a normalization constant, and will provide the same performance. Usually $\kappa = 1/(\mathbf{d} - \mathbf{m})^T \mathbf{K}^{-1}(\mathbf{d} - \mathbf{m})$ minimizes the output variance subject to $\mathbf{w}_{MF}^T \mathbf{d} = 1$. In this paper, $\kappa$ is set in this way. The output value of the MF detector for the $i$th pixel can be computed by

$$y_i = \mathbf{w}_{MF}^T \mathbf{r}_i, \quad 1 \leq i \leq N. \tag{2}$$

**2.2 Relationship between the MF detector and the band significance**

The data matrix $\mathbf{R}$ in (1) is defined as a collection of $N$ pixels, but it can also be expressed as a collection of $L$ bands, which is given by

$$\mathbf{R} = \begin{bmatrix} R_1 \\ R_2 \\ \vdots \\ R_L \end{bmatrix}, \quad (3)$$

where $R_j$ represents the $j$th band image, $1 \leq j \leq L$. Accordingly, the MF output result in (2) can been rewritten in the matrix form as follows:

$$Y = \mathbf{w}_{MF}^T \mathbf{R} = w_1 R_1 + w_2 R_2 + ... + w_L R_L. \quad (4)$$

The expression in (4) indicates that the output image of MF is the weighted sum of all $L$ bands. The absolute value of the $j$th weight, $w_j$ represents the significance of the $j$th band for the target to distinguish from the background, $1 \leq j \leq L$.

An example is presented in the following for illustration. Figure 1 shows the 3 band images of a simulated $51 \times 51 \times 3$ dataset. Each band follows a standard normal distribution. The central $3 \times 3$ region is set as the target pixel area with the value for the first and third band set to 255. According to (1), we can have the weight coefficient of the MF detector as $\mathbf{w}_{MF}$ = [1.4500, 0.0466, 1.5622]$^T$. Obviously, the MF weight of the second band is much lower than those of the other two bands. This is because the second band provides no useful information for distinguishing target pixels from background ones. The MF detector gives almost the same weight to the first and third bands, because they play almost the same role for the detection of targets.

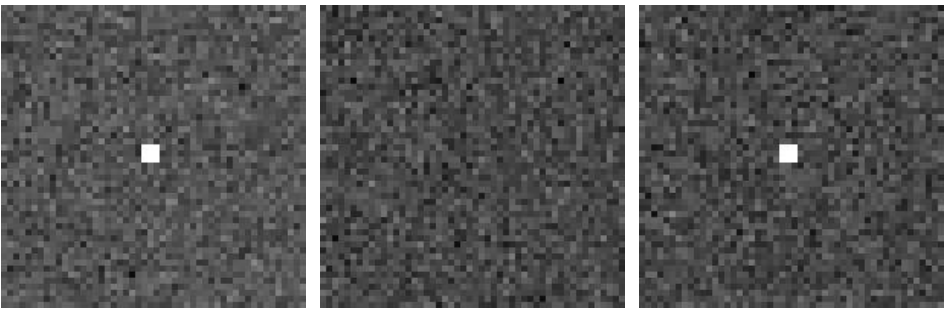

Figure 1, Three bands of the simulated data: (a) Band 1, (b) Band 2 and (c) Band 3. The middle points are the target area.

Therefore, the $L$ MF weight coefficients in (1) reflect the role of each band in helping the target to be distinguished from the background. In other words, the MF has the ability to automatically identify the important bands, and also the bad bands for the target.

**2.3 An automatic bad band preremoval method**

For a given target, the MF weight is directly related to the band importance. However, another issue which we should notice here is that $j$th MF weight, $w_j$ is also affected by the norm of the $j$th band. First, we will give a general formula of the MF detector when the data matrix is transformed by an arbitrary invertible matrix. Suppose the transformation matrix is $\mathbf{A}_{L \times L}$, the transformed data set $\tilde{\mathbf{R}}$ can be computed by

$$\tilde{\mathbf{R}} = \mathbf{A}\mathbf{R}. \tag{5}$$

Then, the new MF detector, $\tilde{\mathbf{w}}_{\text{MF}}$ and output image $\tilde{Y}$ can be calculated according to (1), (4) and (5) respectively,

$$\tilde{\mathbf{w}}_{\text{MF}} = \left(\mathbf{A}^T\right)^{-1} \mathbf{w}_{\text{MF}}, \tag{6}$$

$$\tilde{Y} = \tilde{\mathbf{w}}_{\text{MF}}^T \tilde{\mathbf{R}} = \mathbf{w}_{\text{MF}}^T \mathbf{A}^{-1} \mathbf{A} \mathbf{R} = Y. \tag{7}$$

That is to say, the linear transformation on the data set $\mathbf{R}$ has no impact on the final MF output $Y$, as illustrated in (7). However, the situation for the MF detector is different, as shown in (6). A simple example to understand (6) is that if we double all values in the $j$th band, the corresponding $j$th MF weight will be only one-half of the original one. In other words, if a noisy band has a very low value relative to other bands, its absolute value of MF weight could be larger than those of the other bands.

Therefore, the normalization of each band is required before applying MF weights for bad band removal. Let the transformation matrix $\mathbf{A}$ be

$$\mathbf{A} = \begin{bmatrix} \|R_1\| & 0 & \cdots & 0 \\ 0 & \|R_2\| & \cdots & 0 \\ \vdots & \vdots & \ddots & \vdots \\ 0 & 0 & \cdots & \|R_L\| \end{bmatrix}, \tag{8}$$

which is an *L*-dimensional diagonal matrix with the (*j*, *j*) element corresponding to the norm of Band *j* (*j*=1,2,…,*L*). Obviously, $\mathbf{A}^T=\mathbf{A}$. And the normalized MF (NMF) detector can be computed by combining (1) and (6) as

$$\mathbf{w}_{NMF} = \kappa \mathbf{A}^{-1} \mathbf{K}^{-1} (\mathbf{d} - \mathbf{m}). \tag{9}$$

Now, the NMF weight can really represent the band importance to the target **d** without being influenced by the norm of each band. However, besides band norm, the MF or NMF weights are also the target **d** itself. That is, for the same dataset, the NMF weights vary if different targets are selected. Therefore, different bad band ranges may be generated for different targets. However, their intersection should be the real bad bands. This is because these bands are of little use to detect any of the selected targets.

The algorithm for using the NMF coefficient to remove bad bands is summarized in Algorithm 1. We first randomly select *M* target pixels in the image, and then compute their mean absolute value (MAV) of NMF weights. By a manually set threshold, *thres*, we can find the bad bands. It should be noted here that *thres* cannot be automatically estimated currently. Generally, *thres* is determined by the sensor, image scene and data quality.

**Algorithm 1: Automatic Bad Band preremoval algorithm**

**Input:**  The data set $\mathbf{R}_{L \times N}$, the number of targets selected, *M*, and the MF threshold, *thres*;

**Step 1:**  Band centralization. Compute the mean vector $\mathbf{m} = [m_1, m_2, \cdots, m_L]$. And for Band *j*, do $R_j = R_j - m_j$, (1<*j*<*L*);

**Step 2:** Compute the covariance matrix, $\mathbf{K} = \mathbf{R}\mathbf{R}^T / N$, and the transformation matrix, **A** as (8);

**Step3:** Randomly select $M$ pixels as targets, and their indices are noted as $\{\text{ind}_s\}$, $(1 \leq s \leq M)$. For the $s^{\text{th}}$ target, let $\mathbf{d}_s = [\mathbf{R}]_{:,\text{ind}_s}$, and compute the NMF weight vector as (9);

**Step4:** Compute the mean absolute value (MAV) of the NMF coefficient as

$$\overline{\mathbf{w}}_{\text{NMF}} = mean\left(\left\{abs\left(\mathbf{w}_{\text{NMF}}^1\right), abs\left(\mathbf{w}_{\text{NMF}}^2\right), ..., abs\left(\mathbf{w}_{\text{NMF}}^M\right)\right\}\right); \tag{10}$$

**Output:** Band numbers with the MAV of NMF weights no larger than *thres*.

## 3, Experiments

In this section, three commonly used real hyperspectral images (downloaded at http://www.ehu.eus/ccwintco/index.php?title=Hyperspectral_Remote_Sensing_Scenes) are applied to test the performance of our algorithm. In previous studies, the bad bands of the three data sets were manually removed by the researchers. For the three data sets, we mark their reference bad band ranges in figure 2, 5, and 7 respectively. To illustrate how the selected bands vary with *thres*, two values of *thres* are selected for each data set.

### 3.1 Indian Pines Data

The data set used in this experiment is the frequently utilized Indian Pines data(Baumgardner, Biehl, and Landgrebe 2015). This scene was gathered by the AVIRIS hyperspectral sensor on June 12, 1992 and consists of 145×145 pixels and 220 spectral reflectance bands in the wavelength range 0.4-2.5 μm. The main land cover type of the Indian Pines scene is vegetation, which contains two-thirds agriculture, and one-third forest or other natural perennial vegetation. In previous research, the identified noisy bands included Band 1-3,103-112,148-165, and 217-220, which were corrupted due to water absorption or instrumental problems (Kang, Xiurui, and Luyan 2015; Archibald and Fann 2007; Kang et al. 2014). These bands are usually manually removed first in the data preprocessing step.

In this and the following experiments, we purposefully keep these bands to verify the ability of our algorithm to identify them.

First, we set the number of selected target pixels as $M=1000$, and the MAV of NMF weight coefficients as a function of the band index is shown in figure 2. It can be seen that low NMF weights generally correspond to the reference noisy bands. If we set *thres*=1 and 1.5, the corresponding bad bands selected by NMF are listed in table 1. More bands are chosen when *thres* is set with a larger value. Water absorption bands, like Band 103-109, 149-164, 218-220 and the low-SNR bands, like Band 1-2, are selected. Some images of those bands are shown in figure 3 (a-c). It can be seen that they indeed contain no useful information. However, we can find that Band 61-62, 75-76, 83-97, which do not fall in the range of the reference noisy bands, are also selected. The images of Band 61, 75 and 90 are demonstrated in figure 3 (d-f), which are all corrupted bands. On the contrary, some reference noisy bands, like Band 3, 110-112, 148, 165, 217, are not identified as bad bands by our algorithm. We draw the images of Band 3, 112, 217 in figure 3 (g-i). Though noise exists in these bands, the distribution of different ground objects can been distinguished generally. As a result, they have a higher MF weight value than the previous bands.

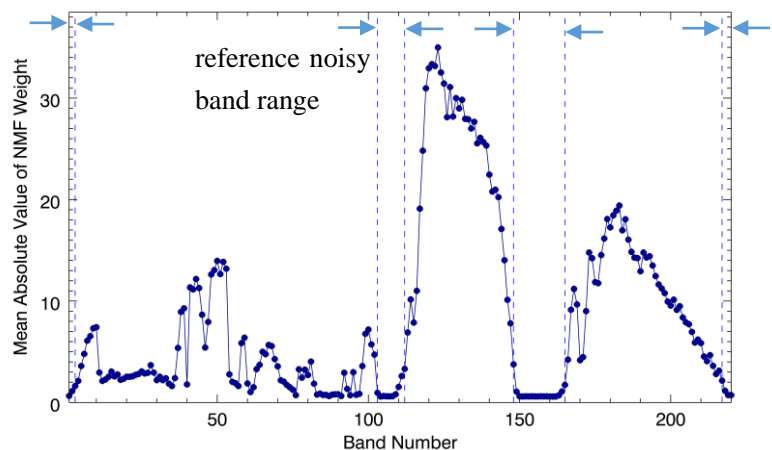

Figure 2, the spectrum of the MAV of the NMF weight for the Indian Pines data.

Table 1, Band indices with corresponding MAV of the NMF weight no larger than 1 and 1.5 for the Indian Pines data.

| NMF *thres* | Number of selected bands | Band No. |
|---|---|---|
| 1 | 37 | 1, 76, 83-97, 103-109, 150-163, 219-220 |
| 1.5 | 45 | 1-2, 61-62, 75-76, 83-97, 103-109, 149-164, 218-220 |

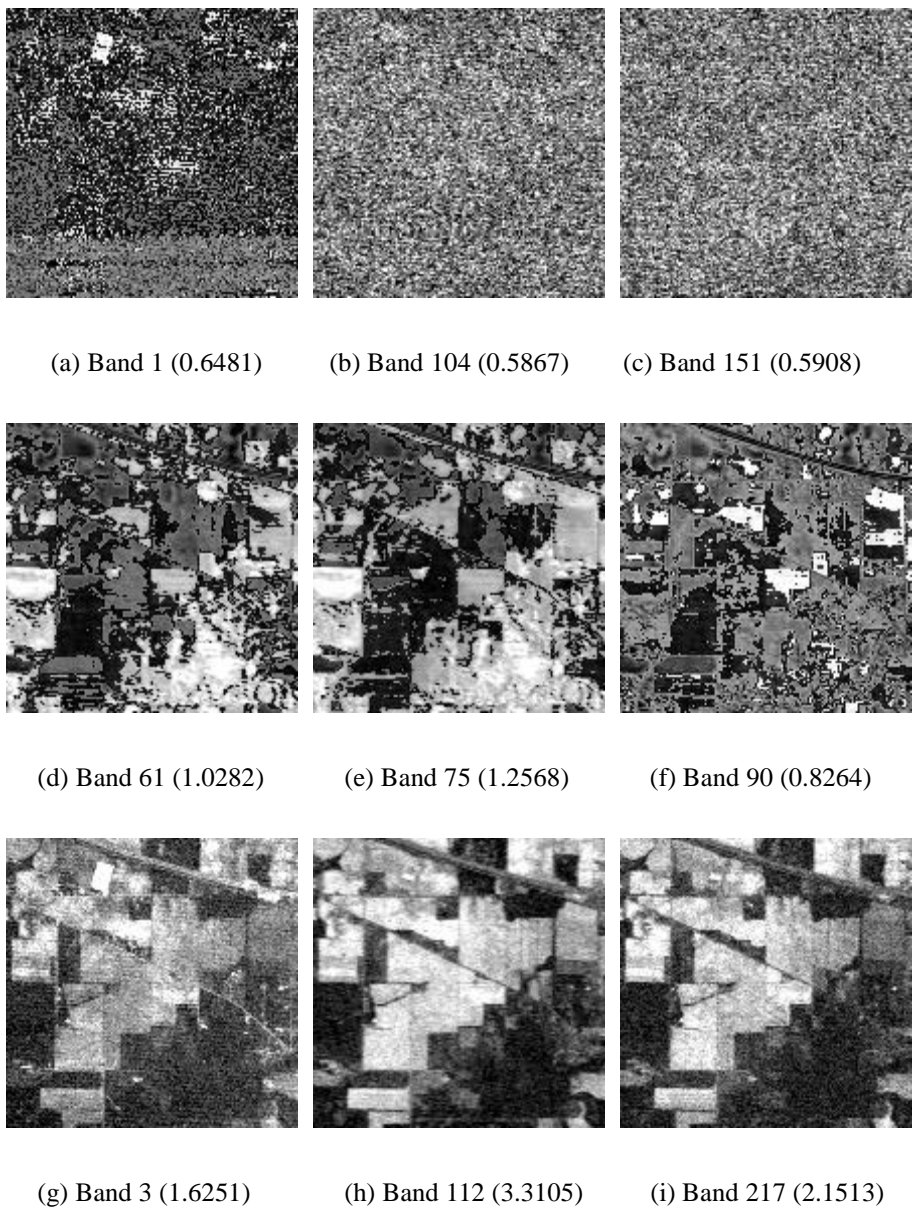

(a) Band 1 (0.6481)   (b) Band 104 (0.5867)   (c) Band 151 (0.5908)

(d) Band 61 (1.0282)   (e) Band 75 (1.2568)   (f) Band 90 (0.8264)

(g) Band 3 (1.6251)   (h) Band 112 (3.3105)   (i) Band 217 (2.1513)

Figure 3, the greyscale images for Band 1 (a), 104 (b), 151 (c), 61 (d), 75(e), 90(f), 3 (g), 112 (h) and 217 (i) of the Indian Pines data. The value in parentheses corresponds to the MAV of the NMF weight.

In the above experiment, the number of selected target pixels, M is manually set to M =1000. Next, we will conduct the sensitivity analysis of our algorithm to M. In this experiment, M is set from 1 to 10000, with an interval of 1 in the range [1, 10], 10 in [10,100], 100 in [100,1000] and 1000 in [1000,10000]. Again, two MF thresholds are selected, i.e. *thres*=1 and 1.5. Since the target pixels are randomly selected, we repeat 20 runs for each M. The final number of selected bands related to M is plotted in figure 4. And we can see that for both thresholds, the number of selected bands declines dramatically with M when M is small, and converges to a steady level when M is large. The values of M when the number of selected bands begin to be stable are M = 70 (0.33% of all pixels) and 600 (2.85% of all pixels) for the thresholds, 1 and 1.5 respectively. That is to say, only a small portion of pixels are required for our algorithm to determine the bad bands.

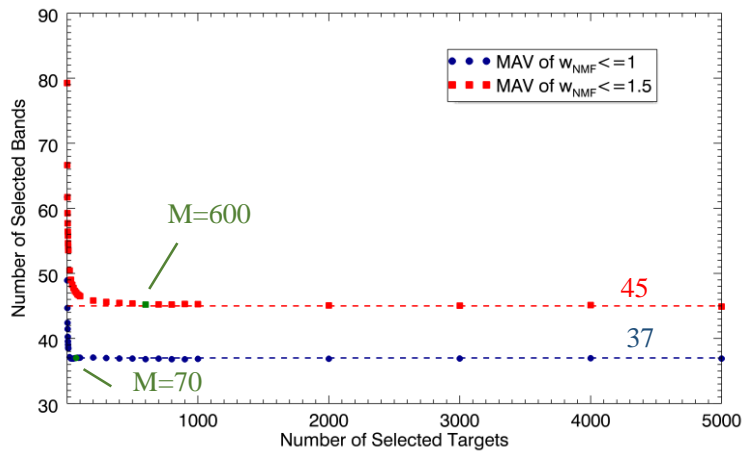

Figure 4, the number of selected bands as a function of the number of selected targets with the MAV of the NMF weight threshold set to 1 and 1.5.

**3.2 Salinas Data**

In this experiment, another AVIRIS hyperspectral dataset is used, which was collected over the Salinas Valley, California, and is characterized by high spatial resolution (3.7-meter pixels). The area covered comprises 512 lines by 217 samples with 224 bands. In previous studies, the reference noisy

bands are Band 1-3,106-114,150-167,221-224 (Kang, Xiurui, and Luyan 2015; Makarau et al. 2012; Mirzapour and Ghassemian 2015), and remain in this study. As before, the number of selected target pixels is set to *M*=1000. The spectrum of the average MF weight is shown in figure 5. Generally speaking, the weight value of bands in the reference noisy bands is lower than that of the other bands.

The bands with MF weight no larger than 5 and 10 are summarized in table 2. A total of 15 and 29 bands are identified as bad bands respectively. Except for Band 4, the identified bands all belong to the reference band range. Three representative bands, Band 1, 110 and 159 are shown in figure 6 (a-c) and it can be seen that their data quality is poor. Comparatively, the data quality of Band 4 (figure 6(d)) is much better than the above 3 bands, but still has a relatively high noisy level. As in the above experiment, some reference noisy bands, like Band 106, 114, 150-152,167, are not selected as bad bands by our algorithm. Band 106 and 150 are shown in figure 6 (e-f). Clearly, these two bands have a high data quality and are suitable for further data processing.

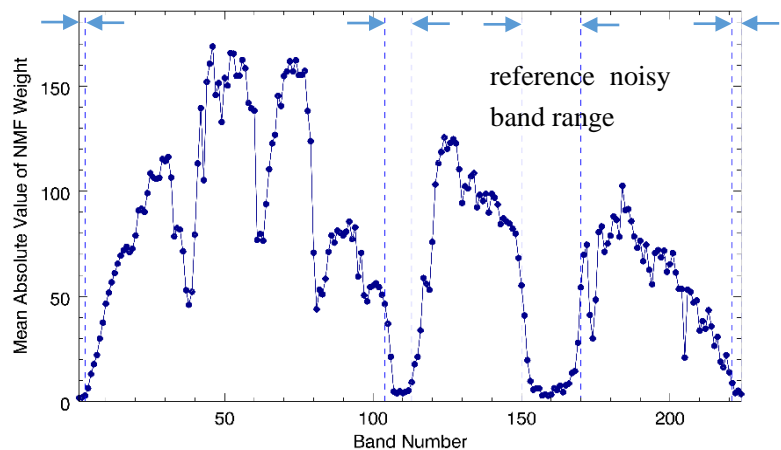

Figure 5, the spectrum of the MAV of the NMF weight for the Salinas data.

Table 2, Band indices with corresponding MAV of the NMF weight no larger than 5 and 10 for the Salinas data.

| NMF *thres* | Number of selected bands | Band No. |
|---|---|---|
| 5 | 15 | 1-3, 107-111, 157-164, 222, 224 |
| 10 | 29 | 1-4, 107-113, 153-166, 221-224 |

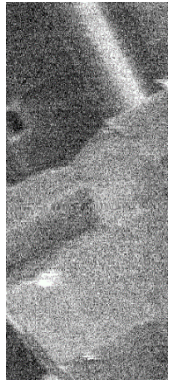 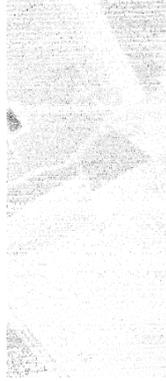 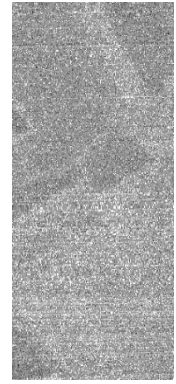

(a) Band 1 (1.7680)  (b) Band 110 (3.9593)  (c) Band 159 (2.7421)

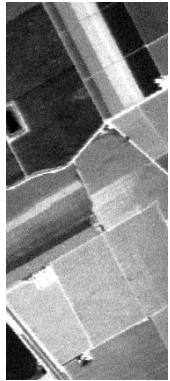 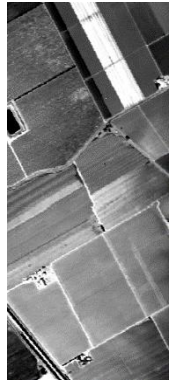 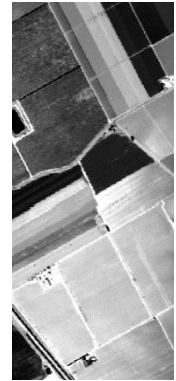

(d) Band 4 (6.3612)  (e) Band 106 (21.2376)  (f) Band 150 (55.1478)

Figure 6, the greyscale images for Band 1 (a), 110 (b), 159 (c), 4 (d), 106 (e) and 150 (f) of the Salinas data. The value in parentheses corresponds to the MAV of the NMF weight.

### 3.3 Cuprite Data

In this experiment, the well known Cuprite data (f970619t01p02_r02_sc03.a.rfl reflectance file) with a size of 512 × 614 × 224 is used. In previous studies, the reference noisy bands are Band 1-3,104-115,150-170,221-224 (Kang, Xiurui, and Luyan 2015; Martń and Plaza 2010; Iordache,

Bioucas-Dias, and Plaza 2011; Plaza et al. 2009). Within these bands, Band 1-2, 108-111, 115-116 are all zero bands. In order to apply (1), those bands are reset with random noise. As before, the number of selected target pixels is set to $M = 1000$. The spectrum of the average MF weight is shown in figure 6, where the red circles represent zero bands. Band numbers with MF weight coefficients no larger than 5 and 10 are summarized in table 3. The MF weight coefficients of 67.5% reference nosiy bands are no larger than 5, while the MF coefficients of 87.5% reference noisy bands are no larger than 10. Two representative bands (Band 3 and 112) are shown in figure 8 (a-b). However, 12.5% reference noisy bands should not refereced as noisy bands, as Band 104 and 150 in figure 8 (c-d). Conversely, some bands that are not in the reference noisy band range also have a smaller MF weight, such as Band 114 and 220 in figure 8(e-f). Compared to Band 104 and 150, the noisy levels of Band 114 and 220 are higher.

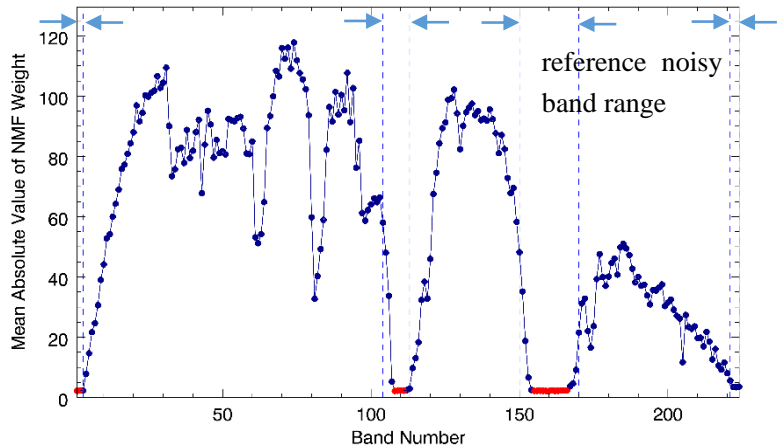

Figure 7, the spectrum of the MAV of the NMF weight for the Cuprite data. Bands with red circles correspond to zero bands.

Table 3, Band indices with corresponding MAV of the NMF weight no larger than 5 and 10 for the Cuprite data.

| NMF *thres* | Number of selected bands | Band No. |
| --- | --- | --- |

| 5 | 27 | 1-3,108-113,154-168,222-224 |
| 10 | 41 | 1-4,107-114,153-169,218,220-224 |

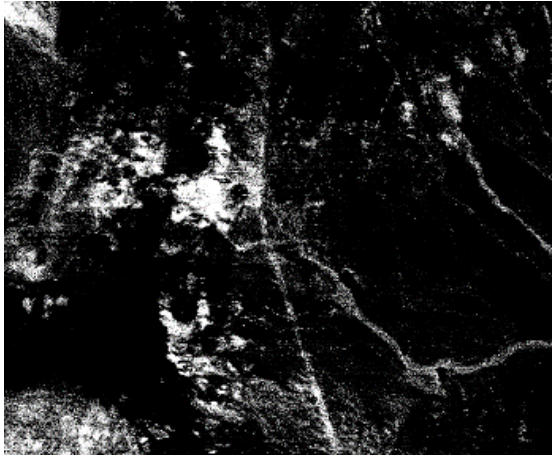

(a) Band3 (2.3283)

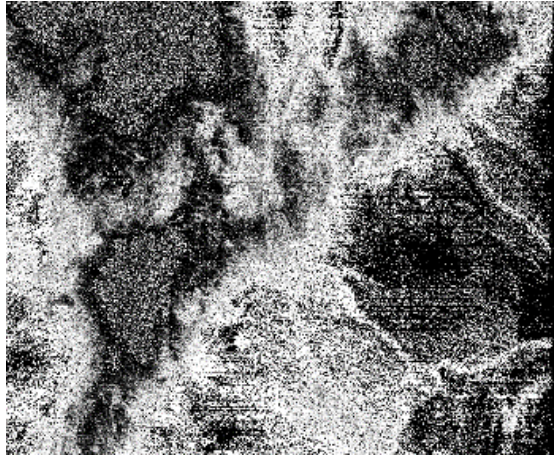

(b) Band 112 (2.4278)

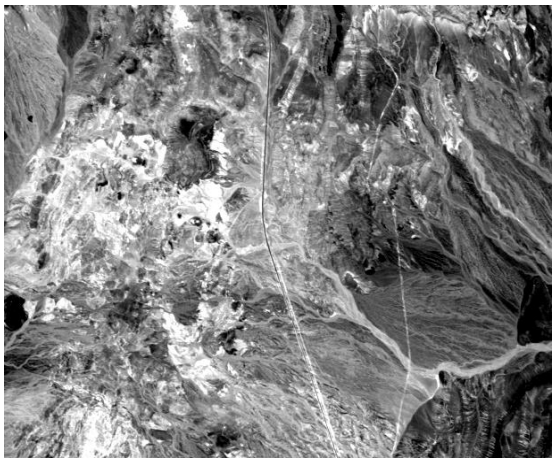

(c) Band 104 (58.0089)

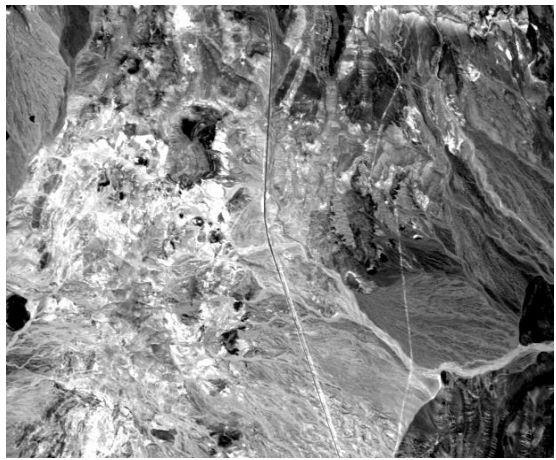

(d) Band 150 (48.1969)

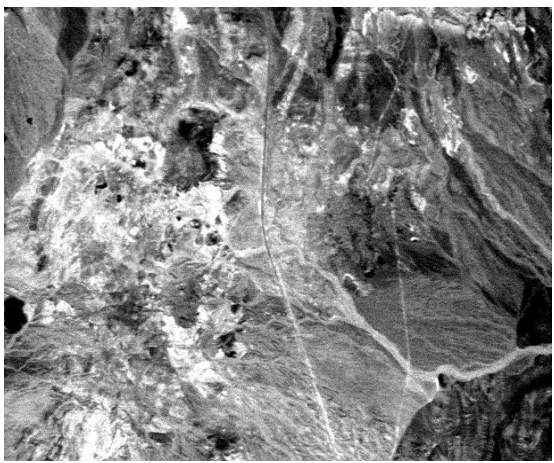

(d) Band 114 (9.7515)

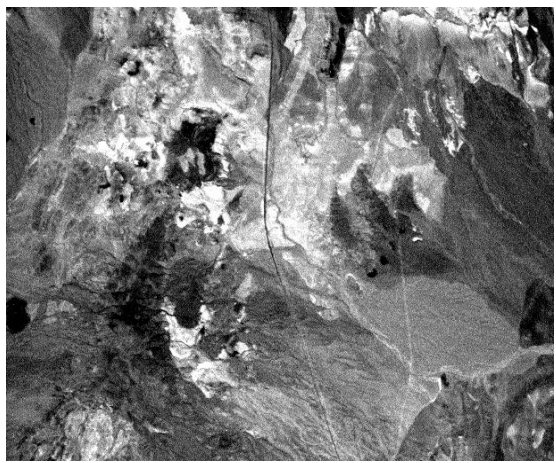

(e) Band 220 (8.0814)

Figure 8, the greyscale images for Band 3 (a), 112 (b), 104 (c), 150 (d), 114 (e) and 220 (f) of the Cuprite data. The value in parentheses corresponds to the MAV of the NMF weight.

**4. Conclusion**

In this paper, we have established the relation between the MF target detection method and the band importance ranking. The fact is that the *L* weights of the MF detector actually indicate the significance of each band in differentiating the target from the background. Accordingly, an automatic algorithm for bad band removal is proposed by utilizing the average weights of *M* randomly selected target pixels. However, since the MF weight of one band is related to the norm of that band, the data should be normalized in each band first. Three widely applied hyperspectral images (Indian Pines, Salinas and Cuprite data) are used to verify the presented method, and the reference bad band ranges are summarized according to previous studies. Experimental results show that our algorithm can find all the low SNR bands in the reference bad band ranges for all the three data sets. On the contrary, bands, which are in the reference bad bands ranges but with a high SNR level, will not be selected by our method. Moreover, we can also identify bad bands not within the reference ranges. Therefore, it is an effective automatic algorithm, and can be used as the very first preprocessing step for remote sensing imagery.